\documentclass[letterpaper]{article} % DO NOT CHANGE THIS
\usepackage{aaai24}  % DO NOT CHANGE THIS
\usepackage{times}  % DO NOT CHANGE THIS
\usepackage{helvet}  % DO NOT CHANGE THIS
\usepackage{courier}  % DO NOT CHANGE THIS
\usepackage[hyphens]{url}  % DO NOT CHANGE THIS
\usepackage{graphicx} % DO NOT CHANGE THIS
\usepackage{natbib}  % DO NOT CHANGE THIS AND DO NOT ADD ANY OPTIONS TO IT
\usepackage{caption} % DO NOT CHANGE THIS AND DO NOT ADD ANY OPTIONS TO IT
\usepackage{algorithm}
\usepackage{algorithmic}
\usepackage{bm}
\usepackage{amsmath}
\usepackage{amssymb}
\usepackage{booktabs}
\usepackage{multirow}
\usepackage{makecell}
\usepackage{subfigure}
\usepackage{color}
\usepackage{float}

\usepackage{newfloat}
\usepackage{listings}
\DeclareCaptionStyle{ruled}{labelfont=normalfont,labelsep=colon,strut=off} % DO NOT CHANGE THIS
\floatstyle{ruled}
\newfloat{listing}{tb}{lst}{}
\floatname{listing}{Listing}
\title{PMET: Precise Model Editing in a Transformer}
\author{
    Xiaopeng Li,
    Shasha Li\thanks{Corresponding Author.},
    Shezheng Song,
    Jing Yang,
    Jun Ma,
    Jie Yu$^{\rm *}$,
}
\affiliations{
    National University of Defense Technology\\
    Changsha, Hunan 410073 China\\
 xiaopeng.lyy@gmail.com, \{shashali, ssz614, yj, majun\}@nudt.edu.cn, yangjing2036@126.com
}

\usepackage{bibentry}
\begin{document}

\maketitle

\begin{abstract}
Model editing techniques modify a minor proportion of knowledge in Large Language Models (LLMs) at a relatively low cost, which have demonstrated notable success. Existing methods assume Transformer Layer (TL) hidden states are values of key-value memories of the Feed-Forward Network (FFN). They usually optimize the TL hidden states to memorize target knowledge and use it to update the weights of the FFN in LLMs. However, the information flow of TL hidden states comes from three parts: Multi-Head Self-Attention (MHSA), FFN, and residual connections. Existing methods neglect the fact that the TL hidden states contains information not specifically required for FFN. Consequently, the performance of model editing decreases. To achieve more precise model editing, we analyze hidden states of MHSA and FFN, finding that MHSA encodes certain general knowledge extraction patterns. This implies that MHSA weights do not require updating when new knowledge is introduced. Based on above findings, we introduce PMET, which simultaneously optimizes Transformer Component (TC, namely MHSA and FFN) hidden states, while only using the optimized TC hidden states of FFN to precisely update FFN weights. Our experiments demonstrate that PMET exhibits state-of-the-art performance on both the \textsc{counterfact} and zsRE datasets. Our ablation experiments substantiate the effectiveness of our enhancements, further reinforcing the finding that the MHSA encodes certain general knowledge extraction patterns and indicating its storage of a small amount of factual knowledge. Our code is available at \url{https://github.com/xpq-tech/PMET}.
\end{abstract}

\section{Introduction}
Large language models (LLMs), as an emerging form of knowledge base \cite{petroni-etal-2019-language,heinzerling-inui-2021-language,cao-etal-2021-knowledgeable}, are extensively employed worldwide, primarily addressing queries through knowledge recall. Nonetheless, these models are often criticized for furnishing erroneous information \cite{Ji_2023_Survey_of_Hallucination,zhao2023survey}. The cost of fine-tuning or training from scratch to correct the minor proportion of erroneous knowledge is frequently deemed impractical. Fortunately, recent model editing techniques have demonstrated the ability to modify minor proportion of knowledge in LLMs with relatively low cost \cite{meng2022massediting,mitchell2022fast,yao2023editing}. Model editing aims to modify the internal knowledge of LLMs without resorting to vanilla training or fine-tuning. The success rates of model editing in edited-knowledge and in knowledge related to the edited-knowledge are assessed separately based on efficacy and generalization, while the preservation of irrelevant edited-knowledge is measured by specificity (also known as locality) \cite{mitchell2022fast}. For the purpose of uniformity, we collectively refer to efficacy and generalization as reliability. Additionally, two metrics are employed to evaluate the generative capacity of the post-edited model: fluency and consistency \cite{Meng2022Locating}. Model editing methods can be classified into two categories based on whether the original model weights are modified: \textit{weight-preserved} and \textit{weight-modified} methods \cite{yao2023editing}. Weight-preserved methods often require additional content, whereas weight-modified methods directly edit model weights without the need for extra content, making them a more lightweight alternative.

\begin{figure*}[t]
  \centering
  \subfigure[Previous optimization-based method.]{\includegraphics[scale=0.5, trim=5 0 -5 0,clip]{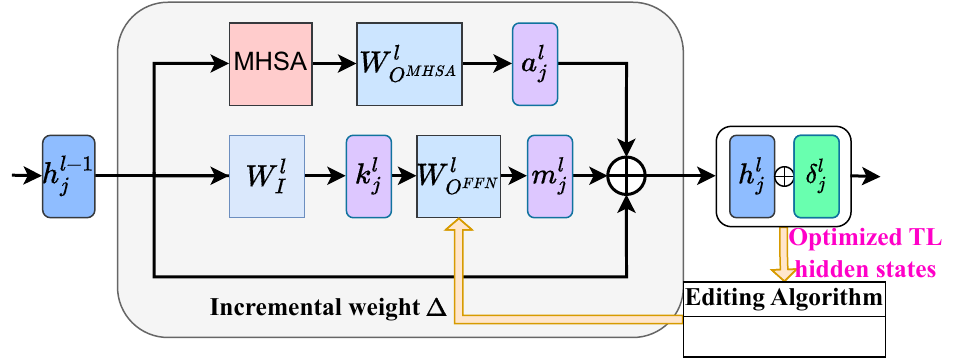}}
  \subfigure[PEMT method.]{\includegraphics[scale=0.5, trim=5 0 -5 0,clip]{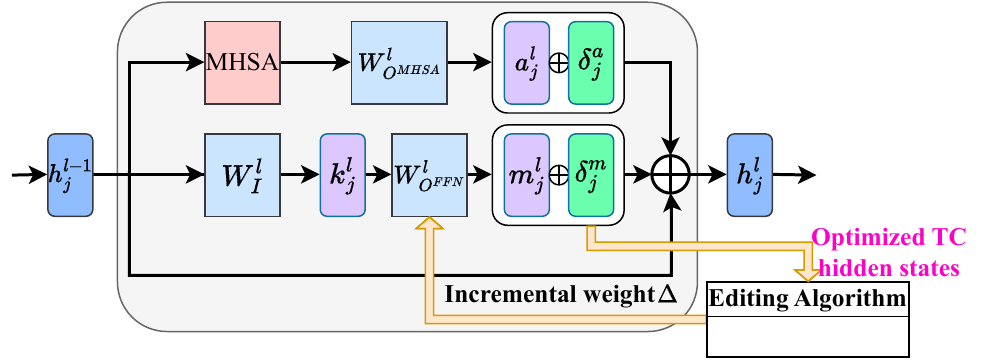}}
  \subfigure{\includegraphics[scale=0.5, trim=0 0 0 0,clip]{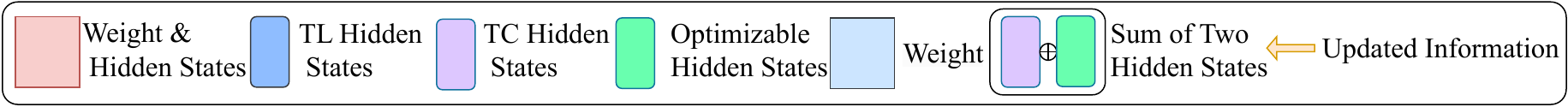}}
  \caption{Comparison between PMET and existing methods in a Transformer layer. (a) Existing optimization-based methods employ optimized TL hidden states to perform vague updates on FFN weights. (b) PMET simultaneously optimizes the TC hidden states of both MHSA and FFN, but only uses the optimized TC hidden states of FFN to perform precise updates on FFN weights.}\label{pmet.comparison}
\end{figure*}
The weight-modified approaches include learning-based \cite{mitchell2022fast} and optimization-based methods \cite{meng2022massediting}. Learning-based methods utilize gradient information to update the weights, but they suffer from poor knowledge generalization and are prone to overfitting. The optimization-based method ROME \cite{Meng2022Locating} alleviates this by solving a optimization problem and updates FFN weights incrementally. The subsequent MEMIT \cite{meng2022massediting} further improves ROME, enabling mass editing in a single operation and demonstrating impressive editing performance. In detail, ROME and MEMIT view FFN as key-value memories \cite{geva-etal-2021-transformer}, where the hidden states before and after passing through FFN weight $W$ can be considered as keys $k$ and values $v$, satisfying $Wk=v$. ROME and MEMIT extract TC (Transformer Component, namely MHSA and FFN) hidden states as keys and optimizes TL (Transformer Layer) hidden states as values, ultimately obtaining the desired weights by solving a least square problem. However, the information flow of TL hidden states comes from three parts: Multi-Head Self-Attention (MHSA), FFN, and residual connections. Both ROME and MEMIT uses optimized TL hidden states as the values (i.e., target knowledge representations) for updating FFN weights, which \textbf{overlooks the irrelevant information within the TL hidden states that is not required by FFN}, resulting in imprecise updating of weights based on non-accurate target knowledge representations and compromising editing performance. To address this issue, we suggest optimizing the TC hidden states directly of FFN to memorize target knowledge for precise updates on FFN weights.

But unexpectedly, during the practical process of directly optimizing the TC hidden states of FFN, we encounter occasional optimization bottlenecks where the TC hidden states can not be aligned with target knowledge. We attribute this phenomenon to the limitations imposed by the parameter space of the TC hidden states of FFN. A plausible solution to address this is the supplementary optimization of TC hidden states within MHSA. However, this introduces a novel inquiry: whether MHSA, like FFN, possesses the capability to store factual knowledge \cite{geva-etal-2021-transformer}, necessitating updates to its weights. In pursuit of this inquiry, we endeavor to analyze hidden states of MHSA and FFN to determine the role of MHSA in LLMs' knowledge recall. We then observe that the knowledge contained within MHSA undergoes more frequent changes compared to that within FFN. Combining previous findings of MHSA \cite{geva2023dissecting,wang2022interpretability,hassid2022much} with our observation, we believe that MHSA works as a knowledge extractor and stores certain general knowledge extraction patterns. This suggests the potential for supplementary optimization of TC hidden states of the MHSA to expand the function space, \textbf{without} necessitating updates to its weights.

Based on the above finding, we propose PMET, which simultaneously optimizes TC hidden states of MHSA and FFN, utilizing solely the optimized TC hidden states of FFN as the target knowledge representations for updating FFN weights, enabling precise updates. The differences between PMET and existing methods are illustrated in Figure \ref{pmet.comparison}. Our experiments demonstrate that PMET exhibits state-of-the-art comprehensive performance in editing GPT-J (6B) \cite{gpt-j} and GPT-NeoX (20B) \cite{black-etal-2022-gpt} on the zsRE and \textsc{counterfact} datasets. Specifically, in \textsc{counterfact} dataset, PMET shows a 3.3\% average reliability enhancement over the state-of-the-art method, while in zsRE dataset, it achieves a 0.4\% average improvement. Furthermore, our series of ablation experiments demonstrate that our enhancements are effective and PMET strikes a good balance among reliability, specificity, fluency, and consistency. To sum up, our main contributions are as follows:
\begin{itemize}
\item We reveal that the MHSA works as a knowledge extractor, encodes certain general knowledge extraction patterns, and stores a small amount of factual knowledge.
\item We propose PMET, which leverages the general knowledge extraction patterns of MHSA and simultaneously optimizes the TC hidden states of MHSA and FFN to memorize target knowledge. However, PMET only uses the optimized TC hidden states of FFN to update FFN weights due to the unnecessary updates to MHSA weights.
\item Our experiments with GPT-J (6B) on the zsRE and counterfact datasets highlight PMET's superiority across multiple dimensions. Additionally, editing GPT-NeoX (20B) on the \textsc{counterfact} dataset underscores PMET's superior reliability and consistency over exsiting methods.
\end{itemize}

\section{Related Work}
\subsection{Model Editing}
Model editing is an emerging field in recent years, mainly aimed at mitigating the high cost of model training. Model editing methods can be classified into two categories: weight-modified and weight-preserved\cite{mitchell2022memorybased,zheng2023edit,hern2023inspecting}. Weight-preserved methods typically achieve this preservation by introducing external models \cite{mitchell2022memorybased}, utilizing in-context learning \cite{zheng2023edit}, or altering the LLMs' representation space \cite{hern2023inspecting}. These approaches effectively safeguard non-target knowledge while modifying the target knowledge. However, as the number of knowledge modifications increases, the required additional content also grows substantially. In contrast, weight-modified methods \cite{sinitsin2020editable,mitchell2022fast,meng2022massediting,Meng2022Locating} directly modify the model weights for editing, thereby avoiding the aforementioned content increasing issue. Initially, weight-modified methods use approaches like multi-loss fine-tune \cite{sinitsin2020editable} and constrained fine-tune \cite{zhu2020modifying}. Yet these methods often suffer from overfitting. To address this issue, researchers later proposed meta-learning methods \cite{de-cao-etal-2021-editing,mitchell2022fast} and optimization-based methods \cite{Meng2022Locating}. Nevertheless, as the number of edited-knowledge increases, the efficacy and generalization of these methods deteriorate significantly. This challenge is tackled by MEMIT \cite{meng2022massediting}, which further improves ROME and enable edit a large amount of knowledge in one go. The optimization-based methods are built upon the conclusion that FFN are key-value memories \cite{geva-etal-2021-transformer} and update FFN weights by optimizing the TL hidden states to memorize target knowledge. While these methods have achieved some success in model editing, they confuse the information flow among the MHSA, FFN, and residual connections, leading to a non-accurate updates on FFN weights. In contrast to these methods, PMET simultaneously optimizes the TC hidden states of MHSA and FFN to memorize target knowledge, while only use the optimized TC hidden states of FFN, facilitating precise updates of FFN weights.

\subsection{Post-Hoc Explanation of Transformers}
Post-hoc explanation is a broad field, and our focus is on understanding the roles of the two components, MHSA and FFN, in Transformers \cite{kovaleva-etal-2019-revealing,wang2022interpretability,hassid2022much,geva-etal-2022-transformer,kobayashi2023feed,Hao_Dong_Wei_Xu_2021,geva2023dissecting}.
Currently, it is widely believed that FFN serves as the main carrier for storing factual knowledge\cite{geva-etal-2021-transformer,Meng2022Locating,meng2022massediting}, and each layer of FFN contributes to knowledge recall\cite{geva-etal-2022-transformer,geva2023dissecting}. MHSA is primarily responsible for capturing the degree of association between different tokens, focusing on interactions between content \cite{Hao_Dong_Wei_Xu_2021,kobayashi2023feed}, and extracting attributes of subjects \cite{geva2023dissecting}. Other studies have shown that MHSA contains different levels of redundant information \cite{wang2022interpretability,hassid2022much}. These findings imply that MHSA may store certain general patterns used for knowledge extraction. However, they do not completely clarify this and fail to provide insights into whether MHSA stores factual knowledge. We endeavor to analyze the hidden states of MHSA and FFN to explore these.

\section{Methodology}
\subsection{Preliminaries}\label{sec.prems}
\subsubsection{Language Modeling} We focus on autoregressive, decoder-only LLMs denoted as $\mathcal{F}_{\theta}$. These models transform the input sequence ${\boldsymbol{x}}$ into $z$ tokens $x_1, ..., x_z$ and then input them into $L$ layers of Transformer decoders to obtain the probabilities of the next token $x_{z+1}$ as follows:
\begin{equation}\label{pmet.llms}
\small
\begin{aligned}
  \mathcal{F}_{\theta}(x_1, ...,x_z) &=\text{softmax}\left(W_{\text{E}}\gamma\left(h_z^{L-1}+a_z^L+m_z^L\right)\right)\\&= \mathbb{P}\left(x_{z+1}|x_1, ...,x_z\right)
  \end{aligned}
\end{equation}
Here, $W_E$ and $\gamma$ represent the embedding matrix and layernorm, respectively, and $a_z^L$ and $m_z^L$ are the TC hidden states of the MHSA and FFN of the $L$-th layer, respectively. Note that the MHSA and FFN in \eqref{pmet.llms} are parallel \cite{gpt-j,black-etal-2022-gpt}. The general forms of the MHSA and FFN at the $l$-th layer and the $j$-th token $x_j^l$ are given by:
\begin{equation}\label{pmet.attnmlp}
\small
\begin{aligned}
  a_j^l &= W^l_{O^{\text{MHSA}}}\text{MHSA}^l\left(\gamma\left(h^{l-1}_1,  h^{l-1}_2,...,h^{l-1}_j\right)\right),\\
  m_j^l &=W_{O^{\text{FFN}}}^l\sigma\left(W_I^l\gamma\left(h_j^{l-1}\right)\right)
\end{aligned}
\end{equation}
Here, {$W^l_{O^{\text{MHSA}}}$} and {$W_{O^{\text{FFN}}}^l$} are the output weights of the MHSA and FFN at the $l$-th layer, respectively, and $\sigma$ represents the non-linear activation function. We have omitted bias terms for simplicity.

\subsubsection{Model Editing Problem} Previous researches on model editing have been limited to defining the problem solely based on editing the triples (i.e., subject-relation-object) themselves \cite{Meng2022Locating,mitchell2022memorybased}, overlooking the knowledge contained within the triples. Consequently, the edited models are unable to reason based on the edited knowledge \cite{cohen2023evaluating}. In this paper, we redefine the model editing problem from a subject-centric perspective, where the edited knowledge is associated with the subject, aiming to enable the edited models to reason based on the subject.

Let $\mathcal{F}_{\theta}$ be an LLM that has learned $N$ pieces of knowledge related to subject $S$, represented by the set:
\begin{equation}\label{pmet.knowledge_set}
  K^S = \left\{ \left\{ {\boldsymbol{x}}^S_i,{\boldsymbol{y}}^S_i \right\}, i \in  \left\{0,1,2,...,N\right\} \right\}
\end{equation}
Here, ${\boldsymbol{x}}^S_i$ and ${\boldsymbol{y}}^S_i$ represent the knowledge clue sequence and the knowledge point sequence, respectively, for the $i$-th piece of knowledge. For example, for subject `Shakespeare,' a knowledge clue about the subject could be: ``Shakespeare is a,'' and the knowledge point about the knowledge clue is ``playwright.'' The LLM $\mathcal{F}_{\theta}$ satisfies: {$\mathcal{F} _{\theta}\left({\boldsymbol{x}}^{S}_i\right)={\boldsymbol{y}}^{S}_i, \forall i \in  \left\{0,1,2,...,N\right\} $}. The objective of model editing is to modify $N'$ pieces of knowledge in the LLM related to subject $S$ to the target knowledge:
\begin{equation}\label{pmet.targetk}
  K^{S_t} =\left\{ \left\{ {\boldsymbol{x}}^{S_t}_i,{\boldsymbol{y}}^{S_{t}}_i \right\}, i \in\left\{0,1,2,...,N'\right\}\right\}
\end{equation}
while keeping the $M(M\gg N)$ pieces of knowledge in the set
\begin{equation}
 K^{\neg S_t} = \left\{ \left\{ {\boldsymbol{x}}^{\neg S_t}_j,{\boldsymbol{y}}^{\neg S_{t}}_j \right\}, j \in\left\{0,1,2,...,M\right\}\right\}
\end{equation}
that are unrelated to the $N'$ pieces of model learned knowledge. Hence, the edited LLM $\mathcal{F}_{\theta^*}$ should satisfy:
\begin{equation}\label{pmet.editedllm}
\begin{aligned}
\mathcal{F}_{\theta^*}&\left({\boldsymbol{x}}^{S_t}_i\right)={\boldsymbol{y}}^{S_t}_i \land \mathcal{F}_{\theta^*}\left({\boldsymbol{x}}^{\neg S_t}_j\right)={\boldsymbol{y}}^{\neg S_t}_j,  \\&\forall i\in \left\{0,1,2,...,N'\right\}, j\in  \left\{0,1,2,...,M\right\}.
\end{aligned}
\end{equation}
The evaluation metrics for model editing can be found in Appendix B (Appendix will be found in \cite{li2023pmet}).

\subsection{Investigating the Role of MHSA and FFN in LLMs' Knowledge Recall}
Inspired by Geva et al. \cite{geva2023dissecting}, who analyzed critical subject information by mapping intermediate topic representations to vocabulary tokens, we compare the differences of hidden states $h^{l-1}_{z}$ and $a^{l}_{z}$ of the last token before and after (i.e., input and output) flowing through the $l$-th layer MHSA, both in the vector space and the vocabulary space. Similarly, we perform the same analysis on the hidden states $h^{l-1}_{z}$ and $m^{l}_{z}$ of the last token before and after flowing through the $l$-th layer FFN.

We use 1209 factual statements from \cite{Meng2022Locating} as knowledge queries to explore the knowledge contained within GPT-J (6B). The last token position of each query aggregates the information of the entire query; thus, we consider the hidden state of the last token of each query as a representation of the key knowledge related to that query from the LLMs. We hypothesis \textit{\textbf{a positive correlation between the similarity of hidden states and the consistency of knowledge}} \cite{liang2020knowledge}. To measure the similarity, we calculate the cosine similarity of hidden states and the Jaccard similarity \cite{murphy1996finley} of the mapping to vocabulary tokens. Specifically, we extract the hidden states of the last token before and after each layer of MHSA and FFN, compute their cosine similarity, and obtain the top-$k$ tokens in the vocabulary. Subsequently, we calculate the Jaccard similarity between the top-$k$ tokens of the intermediate states before and after the process. The Jaccard similarity is defined as follows:
\begin{equation}\label{pmet.jaccard-sim}
  J_k\left(T\left(h_1 \right),T\left(h_2 \right)\right) = \frac{\left|T\left(h_1 \right)\cap T\left(h_2 \right) \right| }{\left|T\left(h_1 \right)\cup T\left(h_2 \right) \right| }
\end{equation}where $T(h_1)$ and $T(h_2)$ represent the top-$k$ mappings of the hidden states $h_1$ and $h_2$ on the vocabulary, respectively. We set $k=50$ in our experiments.

\begin{figure}[t]
  \centering
  \includegraphics[scale=0.45, trim=13 13 13 10,clip]{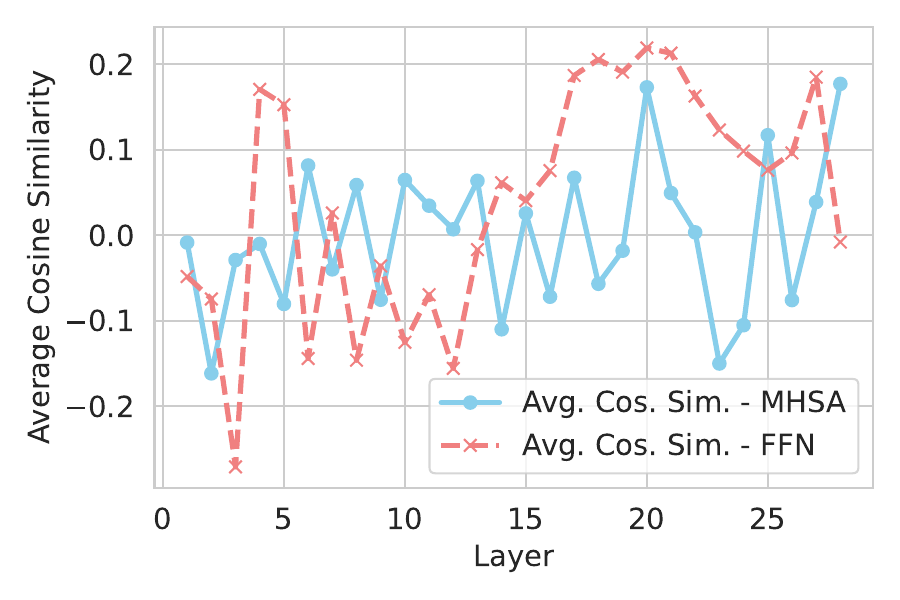}
  \includegraphics[scale=0.45, trim=13 13 13 10,clip]{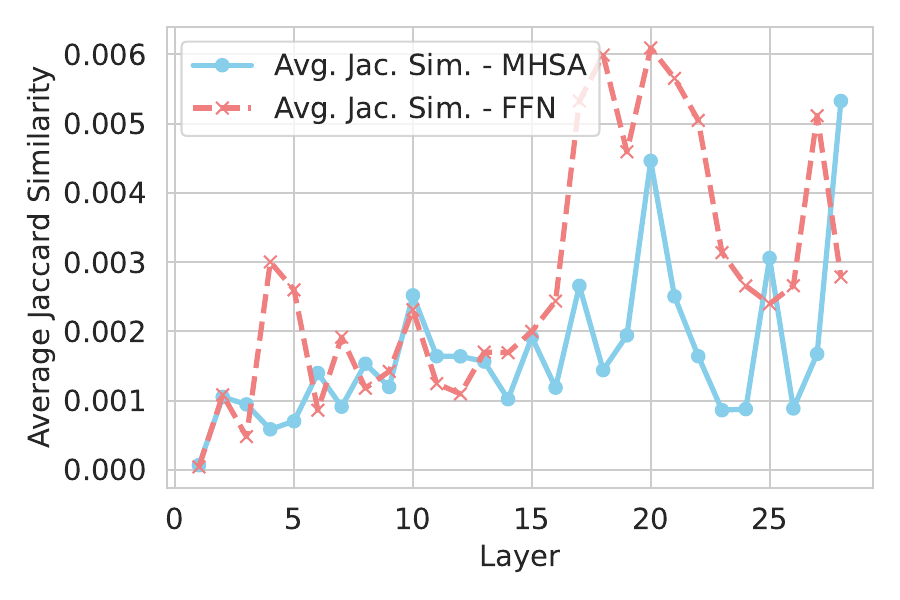}
  \caption{
The changes in the average cosine similarity and average Jaccard similarity of the hidden states before and after MHSA and FFN.}\label{pmet.similarities}
\end{figure}
The average changes in cosine and Jaccard similarities of the last tokens from 1209 factual statements across all layers and components of GPT-J are shown in Figure \ref{pmet.similarities}. \textit{In the first 15 layers of GPT-J, both the MHSA and FFN exhibit relatively frequent changes in their hidden states. However, after the 15th layer, the intermediate states of the FFN undergo a slower rate of change, gradually stabilizing in a specific direction. While the hidden states of the MHSA continue to undergo frequent changes, and their directions remain uncertain throughout the knowledge extraction of GPT-J.} Considering our hypothesis regarding the relationship between hidden states and knowledge, this phenomenon suggests that the knowledge contained in FFN's hidden states tends to become consistent after a certain period, while the knowledge contained in MHSA's hidden states undergoes frequent changes throughout knowledge recall of GPT-J. We attribute this observation to the fact that \textbf{the MHSA continuously extracts various types of knowledge, while the FFN primarily extracts its own knowledge} \cite{geva-etal-2021-transformer,Meng2022Locating}. Furthermore, considering previous findings regarding the extraction of attributes from the MHSA with observed redundancies \cite{geva2023dissecting,wang2022interpretability,hassid2022much}, we believe that the MHSA works as a knowledge extractor and stores certain general knowledge extraction patterns. Thus \textbf{we suggest that when introducing new knowledge, there is no need to update the MHSA weights.}

\subsection{PMET Method}
PMET first computes the target knowledge representations in the last critical layers of FFN by simultaneously optimizing the TC hidden states of both MHSA and FFN. Secondly, PMET only updates FFN weights in the critical layers through target knowledge representations. Overall, PMET optimizes an objective function to obtain target weights \cite{meng2022massediting}:
\begin{equation}
\small
 \begin{aligned}
    W_1 \triangleq \mathop{\text{argmin}}\limits_{{W}} \left( \sum_{i=1}^{n} \left\lVert {W} k_i - v_i \right\rVert^2 + \sum_{i=n+1}^{n+u} \left\lVert {W} k_i - v_i \right\rVert^2 \right).
  \label{pmet.object}
\end{aligned}
\end{equation}

Here, $k_i \triangleq k_i^l$ and $v_i\triangleq v_i^l$ represent the sets of keys and values, respectively, encoding the subject-related knowledge in the $l$-th layer. {$\sum_{i=1}^{n} \left\lVert {W} k_i - v_i \right\rVert^2$} indicates that we want to retain $n$ pieces of knowledge, while {$\sum_{i=n+1}^{n+u} \left\lVert {W} k_i - v_i \right\rVert^2$} indicates that we want to modify $u\gg1$ pieces of knowledge. We represent the keys and values as matrices stacked horizontally: {$\left[ k_1 \mid k_2 \mid \dots \mid k_n \right]\triangleq K $} and {$\left[ v_1 \mid v_2 \mid \dots \mid v_n \right]\triangleq V$}, and we consider the target weight $W_1$ as the sum of the original weight $W_0$ and the incremental weight {$\Delta$}, i.e., {$W_1 = W_0 + \Delta$}. Based on the derivation from MEMIT \cite{meng2022massediting}, the formal expression for the incremental weight is:
{\small \begin{align}\label{pmet.Delta}
  \Delta  & = R K_1^T (C_0 + K_1 K_1^T)^{-1},
\end{align}}where {\small $R \triangleq V_1 - W_0 K_1$} represents the residual between the values $V_1$ (namely target knowledge representations) corresponding to the keys $K_1$ of the target knowledge and the model original knowledge {\small $W_0 K_1$. $C_0 \triangleq K_0 K_0^T = \lambda \mathbb{E}_k\left[ k k^T \right]$} is an estimate of the set of previously memorized keys obtained through sampling. Here, $\lambda$ is a hyperparameter which balances the degree of model modification and preservation.

To explain clearly, let's consider modifying the {$N'$} knowledge instances {$K^S$} related to the subject $S$ in LLMs to the target knowledge {$K^{S_t}$}. Assuming that the set of previously memorized keys {$C_0$} has already been obtained through sampling, and knowledge clues {${\boldsymbol{x}}^{S}_i$} have been inputed into the original model to obtain {$W_0 K_1$}, we then need the sets of keys and values for the target knowledge, denoted as {$K_1$} and {$V_1$}, respectively. Similar to MEMIT, we calculate the target knowledge set of the last critical layer {$L=\max({\mathcal{R}})$}. Throughout this paper, we mainly use {$h^L_i$, $m^L_i$, $a^L_i$}, and {$\delta_i$} to represent the hidden states of the last tokens of the subject $S$ in the knowledge clues {${\boldsymbol{x}}^{S}_i$}.

Unlike ROME and MEMIT, which add optimizable parameters $\delta_i$ to the TL hidden states {$h^L_i$} at the {$L$}-th layer (as shown in Figure \ref{pmet.comparison} (a)) and obtain the optimized TL hidden states {$v_i = h^L_i + \delta_i$} through gradient descent, PMET adds optimizable parameters {$\delta^a_i$} and {$\delta^m_i$} to the TC hidden states {$a^L_i$} and {$m^L_i$} of the components (i.e., MHSA and FFN) at the {$L$}-th layer, respectively. Then, PMET only retains the optimized TC hidden states of FFN to update FFN weights, denoted as {$v^m_i = m^L_i + \delta^m_i = \mathop{\text{argmin}}\limits_{v^m_i}\mathcal{L}(v^m_i)$} (as shown in Figure \ref{pmet.comparison} (b)). {$\mathcal{L}(v^m_i)$} is defined as follows:
\begin{equation}\label{pmet.compute_t}
\small
\begin{aligned}
\mathcal{L}(v^m_i)&=\mu * D_{\text{KL}}\left(\mathbb{P}_{\mathcal{F^{\dagger}_\theta}}\left[\boldsymbol{y} \mid p'  \right]  \lVert \mathbb{P}_{\mathcal{F_\theta}}\left[\boldsymbol{y} \mid p'  \right]  \right)+\\
 &\varphi * \frac{1}{P} \sum_{j=1}^P -\log\mathbb{P}_{\mathcal{F^{\dagger}_\theta}}\left[{{\boldsymbol{y}}_i^{S_t} \mid {\text{pref}}_j \oplus p({\boldsymbol{x_i}})}\right],
\end{aligned}
\end{equation}
where $\varphi$ and $\mu$ are hyperparameters used to balance reliability and specificity. ${\mathcal{F^{\dagger}_\theta}} \triangleq \mathcal{F_\theta}\left( a^L_i+=\delta^a_i, m^L_i+=\delta^m_i\right)$ represents the optimizable parameters {$\delta^a_i$} and {$\delta^m_i$} are added to the TC hidden states of MHSA and FFN at the $L$-th layer of the model $\mathcal{F_\theta}$, respectively. ${\text{pref}}_j \oplus p({\boldsymbol{x_i}})$ and $p'$ are, as in \cite{Meng2022Locating,meng2022massediting}, prefixes used to enhance the generalization of the target knowledge and the prompt template used for calculating the KL divergence: `\{$S$\} is a'. After calculating the values of all the target knowledge that need to be changed, they can be stacked into a matrix $V_1$.

To edit multiple layers of the model, we need to spread the residual $R$ to all critical layers. MEMIT spreads updates evenly over the range of critical layers {$\mathcal{R}$ as $R^l = \frac{V_1 - W_0 K_1}{L-l+1}$}. In contrast, PMET adopts a square root spread to convey more precise information to critical layers:
{\small \begin{align}\label{pmet.R}
R^l = \frac{V_1 - W_0 K_1}{\sqrt{L-l+1}}.
\end{align}}

Now that we have $C_0$ and $R^l$, the next step is to obtain keys $K_1$. Keys are related to specific weights to be edited and represent the hidden states before entering those specific weights. Similar to \cite{Meng2022Locating,meng2022massediting}, the keys $k^l_i$ at the $l$-th layer are defined as follows:
{ \small \begin{align}\label{pmet.ks}
k^l_i =\frac{1}{P} \sum_{j=1}^P \text{prev}(W, {\text{pref}}_j\oplus S),
\end{align}}where {\small $\text{prev}(W,{\text{pref}}_j\oplus S)$} represents the hidden state of the input {\small ${\text{pref}}_j\oplus S$} before flowing through the weight $W$. If one wants to edit {\small $W_{O^{\text{FFN}}}^l$} in \eqref{pmet.attnmlp}, then {\small $\text{prev}(W_ {O^{\text{FFN}}}^l,x)=\sigma\left(W_I^l\gamma\left(h_j^{l-1}\left(x\right)\right)\right)$}.

With this, PMET follows the same algorithm steps as MEMIT to update FFN weights.

\section{Experiments}
\subsection{Baselines and Datasets}
Our experiments are conducted on GPT-J (6B) and GPT-NeoX (20B).
Our baseline methods include the improved Constrained Fine-Tuning (FT+W) \cite{zhu2020modifying,meng2022massediting}, the learning-based method MEND \cite{mitchell2022fast}, and the optimization-based methods ROME \cite{Meng2022Locating} and MEMIT \cite{meng2022massediting}.
For the datasets, we performed counterfactual update experiments on two datasets: Zero-Shot Relation Extraction (zsRE) \cite{levy-etal-2017-zero} and \textsc{CounterFact} \cite{Meng2022Locating}.
More details about datasets can be found in Appendix C.

% =====
\subsection{Editing Experiments}
The score is the harmonic mean of efficacy, generalization, and specificity, representing the balance between reliability (i.e., efficacy and generalization)  and specificity of the editing method\cite{Meng2022Locating}.
Note that in our experiments, we updates counterfactual information, so we evaluated specificity based on factual information, while when testing for efficacy and generalization, we used counterfactual information as the standard.
As a result, the unedited LLMs performed poorly in terms of efficacy and generalization but exhibited good performance in terms of specificity.
Implementation details are presented in Appendix D.
\begin{table*}[t]
  \centering
   \begin{tabular}{lrrrrrr}
    \toprule
        {\textbf{Editor}} & \multicolumn{1}{c}{\textbf{Score}} & \multicolumn{1}{c}{\textbf{Efficacy}} & \multicolumn{1}{c}{\textbf{Generalization}} & \multicolumn{1}{c}{\textbf{Specificity}} & \multicolumn{1}{c}{\textbf{Fluency}} & \multicolumn{1}{c}{\textbf{Consistency}} \\
        \midrule
GPT-J (6B) & 22.4 & 15.2 (0.7) & 17.7 (0.6) & 83.5 (0.5)  & 622.4 (0.3) & 29.4 (0.2)\\\midrule
FT-W & 67.6 & {99.4 (0.1)} & 77.0 (0.7) & {46.9 (0.6)} & {293.9 (2.4)} & {15.9 (0.3)}\\
MEND & {23.1} & {15.7 (0.7)} & {18.5 (0.7)} & {$\boldsymbol{83.0}$ (0.5)}  & 618.4 (0.3) & 31.1 (0.2)\\
ROME & 50.3 & {50.2 (1.0)} & {50.4 (0.8)}  & 50.2 (0.6)  & {589.6 (0.5)} & {3.3 (0.0)}\\
MEMIT & {85.8} & 98.9 (0.2)  & {88.6 (0.5)}  & 73.7 (0.5)  & {619.9 (0.3)} & {40.1 (0.2)}\\
PMET & $\boldsymbol{86.2}$ & $\boldsymbol{99.5}$ (0.1)  & {$\boldsymbol{92.8}$ (0.4)}  & 71.4 (0.5)  & {$\boldsymbol{620.0}$ (0.3)} & {$\boldsymbol{40.6}$ (0.2)}\\
\midrule\midrule
GPT-NeoX (20B) &23.7 & 16.8 (1.9) & 18.3 (1.7)  & 81.6 (1.3) & 620.4 (0.6) & 29.3 (0.5)\\\midrule
MEMIT & 82.0 & 97.2 (0.8)  & 82.2 (1.6) & $\boldsymbol{70.8}$ (1.4)  & $\boldsymbol{606.4}$ (1.0) & 36.9 (0.6)\\
PMET &$\boldsymbol{84.3}$ & $\boldsymbol{98.4}$ (0.2)  & $\boldsymbol{89.4}$ (0.5) & 70.3 (0.5)  & 598.1 (0.6) & $\boldsymbol{38.9}$ (0.2)\\
    \bottomrule
    \end{tabular}
  \caption{10,000 counterfact edits on GPT-J (6B) and GPT-NeoX (20B). Within parentheses is the 95\% confidence interval.}\label{pmet.table-cf}
\end{table*}
\subsubsection{Editing Knowledge in Counterfact}
\begin{figure}[t]
  \centering
  \includegraphics[scale=0.65, trim=3.8 0.9 2.8 0,clip]{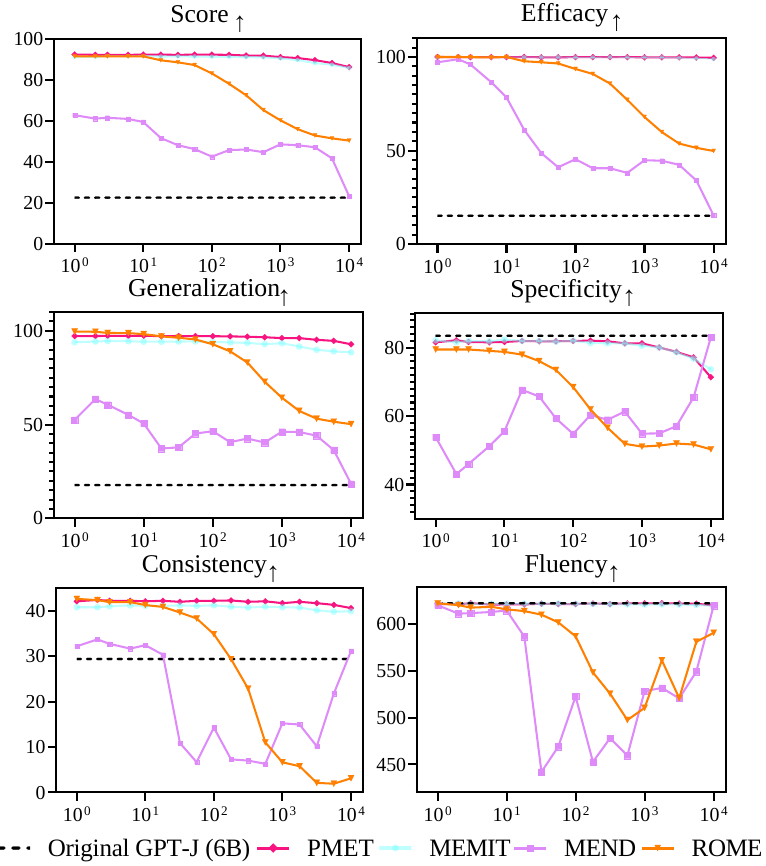}
  \caption{The editing performance of PMET and baselines varies with the number of edits (X-axis).}\label{pmet.scale}
\end{figure}
As mentioned in \cite{Meng2022Locating}, we also conducted 17 counterfactual experiments by sampling 17 integers {\small$n_i=\exp \left(\ln\left(10000\right)*\frac{i}{16}\right)$} from a log-scale curve for editing.
The performance of PMET and other existing methods on GPT-J (6B) in these 17 edits is shown in Figure \ref{pmet.scale}.
With the exception of being slightly inferior to MEMIT in terms of specificity, PMET outperforms all baselines in all other metrics.

Table \ref{pmet.table-cf} shows the results of all methods on 10K counterfactual edits.
The results show that PMET outperforms existing methods in score, efficacy, fluency, and consistency, but is slightly inferior to MEMIT in specificity, and like MEMIT, it is far behind the meta-learning based method MEND.
In the trade-off between editing reliability and specificity, both PMET and MEMIT tend to prioritize reliability, while MEND leans towards specificity.
While sacrificing some specificity for improved reliability is acceptable until better methods are available, we hope to find a compromise in the future.

Then, we applied PMET to conduct 10K edits on GPT-NeoX (20B) on the \textsc{counterfact} dataset, and the results are shown in the lower part of Table \ref{pmet.table-cf}.
Similarly, PMET outperforms MEMIT in terms of reliability and consistency, but lags behind in specificity.
These might be because PMET employs square root propagation \eqref{pmet.R}, resulting in greater changes to the model and hence more damage to specificity .
We further investigate this in the following ablation experiments.
Nevertheless, these results demonstrate that PMET achieves the most significant updates to target knowledge compared to existing methods.
\subsubsection{Editing 10K Knowledge in ZsRE}
The results of editing 10K knowledge on the zsRE dataset are presented in Table \ref{pmet.table-zsre}. The results demonstrate that PMET outperforms existing methods in all three metrics: efficacy, generalization, and specificity. It is worth noting that the original GPT-J (6B) model has a specificity score of only 27.0, and therefore, the specificity of the edited models is also lower than this value.
% =======

\begin{table}[t]
  \centering
\begin{tabular}{lrrrr}
        \toprule
        \textbf{Editor} & \textbf{Efficacy} & \textbf{Generalization} & \textbf{Specificity} \\
        \midrule
GPT-J & 26.4 ($\pm$0.6) & 25.8 ($\pm$0.5) & 27.0 ($\pm$0.5)\\\midrule
FT-W & 69.6 ($\pm$0.6) & 64.8 ($\pm$0.6) & 24.1 ($\pm$0.5)\\
MEND &  {19.4 ($\pm$0.5)} & {18.6 ($\pm$0.5)} & 22.4 ($\pm$0.5)\\
ROME &  {21.0 ($\pm$0.7)} & {19.6 ($\pm$0.7)} & {0.9 ($\pm$0.1)}\\
MEMIT &  {96.7 ($\pm$0.3)} & {89.7 ($\pm$0.5)} & {26.6 ($\pm$0.5)}\\
PMET &  {$\boldsymbol{96.9 }$($\pm$0.3)} & {$\boldsymbol{90.6}$ ($\pm$0.2)} & {$\boldsymbol{26.7}$ ($\pm$0.2)}\\
        \bottomrule
    \end{tabular}
  \caption{10,000 zsRE Edits on GPT-J (6B).}\label{pmet.table-zsre}
\end{table}
% ===========
% ====================  table 3
\begin{table*}[t]
  \centering
     \begin{tabular}{lcllllll}
    \toprule
        \multicolumn{1}{c}{\textbf{Edits}}&\multicolumn{1}{c}{\textbf{Editor}} & \multicolumn{1}{c}{\textbf{Score} } & \multicolumn{1}{c}{\textbf{Efficacy} } & \multicolumn{1}{c}{\textbf{Generalization} } & \multicolumn{1}{c}{\textbf{Specificity} } & \multicolumn{1}{c}{\textbf{Fluency} } & \multicolumn{1}{c}{\textbf{Consistency} } \\
        \midrule
&GPT-J & 22.4 & 15.2 & 17.7 & 83.5  & 622.4 & 29.4 \\  \midrule
\multirow{4}{*}{1K}&PMET & {91.1} &{99.8}  & {{96.1} }  & 80.1   &{{622.2} } & {{41.7} }\\
&\makecell{w/o $\delta^a_i$} & {90.8} ($\downarrow$0.3) & {99.6} ($\downarrow$0.2) & {{96.6} ($\uparrow$0.5) }  & 79.1 ($\downarrow$1.0)  & {{622.4 ($\uparrow$0.2)} }&{{42.2 ($\uparrow$0.5)} }\\
&\makecell{w/ $W^l_{O^{\text{MHSA}}}$} & {90.8 ($\downarrow$0.3)} &{99.8 ($\rightarrow$)} & {{96.8 ($\uparrow$0.7)} } &78.9 ($\downarrow$1.2)  &{{622.0} ($\downarrow$0.2)}&{{42.1 ($\uparrow$0.4)} }\\
&\makecell{Even spread} & {88.9 ($\downarrow$2.2)} &{99.6 ($\downarrow$0.2)}   &{{86.7 ($\downarrow$9.9)} } &82.2 ($\uparrow$ 3.1)  &{{622.3} ($\downarrow$0.1)} &{{39.1 ($\downarrow$3.1)} }\\\midrule
\multirow{4}{*}{5623}&PMET & {88.0}&{99.7} &{{94.5} } & 74.2  &{{621.7} }&{{41.3} }\\
&\makecell{ w/o $\delta^a_i$} &{87.0} ($\downarrow$1.0) &{99.3} ($\downarrow$0.4)  &{{95.0} ($\uparrow$0.5) } & 72.0 ($\downarrow$2.2)  &{{622.3 ($\uparrow$0.6)} }& {{41.6 ($\uparrow$0.3)} }\\
&\makecell{w/ $W^l_{O^{\text{MHSA}}}$} & {86.7} ($\downarrow$1.3) &{99.6} ($\downarrow$0.1) & {{96.0} ($\uparrow$1.5) } &70.8 ($\downarrow$3.4)  &{{621.3 ($\downarrow$0.4)} } &{{41.6 ($\uparrow$0.3)} }\\
&\makecell{Even spread}  &{85.8} ($\downarrow$2.2) &{98.2} ($\downarrow$1.5) &{{82.2} ($\downarrow$12.3) } &79.4 ($\uparrow$5.2)  &{{621.8 ($\uparrow$0.1)} } &{{38.1 ($\downarrow$3.2)} }\\\midrule
\multirow{4}{*}{10K}&PMET & {86.2} & {99.5}  & {{92.8} }  &71.4  & {{620.0} } & {40.6 }\\
&\makecell{w/o $\delta^a_i$} & {85.0 ($\downarrow$1.2)} & {98.9 ($\downarrow$0.6)}   & {{89.0 ($\downarrow$3.8)} } & 71.6 ($\uparrow$0.2)  &{{621.2} ($\uparrow$1.2)} &{{40.0 ($\downarrow$0.6)} }\\
&\makecell{w/ $W^l_{O^{\text{MHSA}}}$}& {84.9 ($\downarrow$1.3)} & {99.5 ($\rightarrow$)}   &{{93.5 ($\uparrow$0.7)} } &68.6 ($\downarrow$2.8)  &{{619.0} ($\downarrow$1.0)} & {{40.5 ($\downarrow$0.1)} }\\
&\makecell{Even spread}  & {83.3 ($\downarrow$2.9)} &{96.7 ($\downarrow$2.8)}   &{{78.4($\downarrow$ 14.4)} } &77.3 ($\uparrow$5.9)  &{{621.8} ($\uparrow$1.8)} &{{37.4 ($\downarrow$3.2)} }\\
    \bottomrule
    \end{tabular}
    
 \caption{The results of the ablation experiments. w/o $\delta^a_i$ represents only optimizing the TC hidden state $\delta^m_i$ of FFN. w/ $W^l_{O^{\text{MHSA}}}$ represents simultaneously updating the weights of both MHSA and FFN. Even spread represents evenly spreading the residual $R$ to the critical layers $\mathcal{R}$.}\label{pmet.ablation}
 \end{table*}
% ===================
\subsection{Ablation Study}\label{pmet.sec.ablation}
We conduct three sets of ablation experiments and demonstrate that:
1) PMET simultaneously optimizing the TC hidden states of MHSA and FFN can result in enhanced reliability; 2) The updating of MHSA weights contributes marginally to the improved generalization of editing while also inflicting greater damage to specificity; and 3) square root spreads in PMET enhances reliability but leads to larger changes in the model, ultimately affecting specificity.
All the ablation experiments were conducted on \textsc{counterfact} using GPT-J (6B), with parameters consistent with the previous experiments in \textsc{counterfact}.

We first conduct experiments where PMET only optimizes TC hidden states of FFN (i.e., $\delta^a_i$ is removed) for 1K, 5623, and 10K counterfactual edits.
The experimental results are shown in Table \ref{pmet.ablation} under ``w/o $\delta^a_i$''.
This shows that simultaneously optimizing TC hidden states of FFN and MHSA can result in better reliability compared to only optimizing TC hidden states of FFN.

Next, we update the weights of both MHSA and FFN using the optimized TC hidden states.
This means we updated {\small$W^l_{O^{\text{MHSA}}}$} and {\small$W^l_{O^{\text{FFN}}}$} simultaneously in \eqref{pmet.attnmlp}, and additionally computed the keys for \eqref{pmet.ks} as {\small$\text{prev}(W^l_{O^{\text{MHSA}}},x)=\text{MHSA}^l\left(h^{l-1}_1\left(x\right),  h^{l-1}_2\left(x\right),...,h^{l-1}_j\left(x\right)\right)$}.
The results are shown in Table \ref{pmet.ablation} under ``w/ $W^l_{O^{\text{MHSA}}}$''.
The results indicate that additionally updating MHSA weights can slightly improve editing generalization, but at the same time, it worsens the specificity.
This might be due to the fact that MHSA weights store certain general knowledge extraction patterns along with a small amount of factual knowledge.
While updating MHSA weights strengthens the extraction patterns of the knowledge similar to edited-knowledge, it may also impair the patterns of extracting other unrelated knowledge, making it more likely to harm specificity.

Finally, we evenly spread the residual $R$ to the critical layers $\mathcal{R}$, and the results are shown in Table \ref{pmet.ablation} under ``Even spread''.
The results indicate that even spreading leads to better model retention (i.e., specificity and fluency), but efficacy, generalization, and consistency are much worse compared to square root spreading.
This suggests that using even spreading in PMET may cause significant loss of update information, reducing the update reliability while preserving more model's original knowledge.
While using square root spreading mitigates the loss of update information, improves reliability, but leads to larger changes in the model, causing more side effects to the specificity and fluency.

We further analyze the relationship between the editing performance and the norms of the incremental weight $\Delta$ in Appendix A. In summary, PMET strikes a good balance between reliability and specificity which becomes more pronounced as the number of edited knowledge increases.

\section{Conclusion}
We reveal that MHSA works as a knowledge extractor and encodes certain general knowledge extraction patterns. Based on this finding, we propose PMET, which simultaneously optimizes the TC hidden states of both MHSA and FFN while only uses the optimized TC hidden states of FFN to perform precise updates of FFN weights. Our experiments on zsRE and \textsc{counterfact} demonstrate the state-of-the-art performance of PMET. Furthermore, our ablation experiments show that our enhancements are effective, PMET strikes a good balance between different metrics, and MHSA stores a small amount of factual knowledge. Our findings contribute additional insights for a better comprehension of the roles played by MHSA and FFN, and our approach takes a step forward in terms of model editing techniques.

\section{Limitations}
Unlike knowledge graphs that explicitly store information in symbolic form \cite{liang2023learn,Liangke}, LLMs implicitly store substantial knowledge in parameterized form. The partial opacity of LLMs' internal mechanisms poses challenges for direct weight modification in model editing. While approaches like PMET and MEMIT have shown promising results in some evaluations, their effectiveness does not necessarily indicate true internalization of edited knowledge by LLMs. Consequently, models edited by PMET and MEMIT cannot reason using the edited knowledge (e.g., after editing the knowledge ``The Prime Minister of the UK is \{Theresa\},'' to ``\{Rishi Sunak\}'' the edited model might generate statements like ``The Prime Minister of England is \{British\}, not \{Indian\}''). Additionally, although this paper defines the problem of knowledge editing centered around subjects, benchmark and dataset construction have not strictly adhered to this definition but instead have been adapted to existing evaluation methods. In the future, we aim to devise more sophisticated editing methods and evaluation metrics (e.g., such as MQuAKE by Zhong et al. (2023) and RIPPLEEDITS by Cohen et al. (2023)) to advance model editing.
\section{Ethical Statement}
The original intention of our research into model editing techniques is to rectify errors and outdated knowledge in LLMs, enabling them to better serve our needs. However, these techniques also have the potential for misuse, allowing LLMs to generate false, toxic, and harmful content. Therefore, we emphasize the importance of not placing excessive trust in the generated content until LLMs are well-regulated.

\section{Acknowledgments}
This work was partly supported by the Hunan Provincial Natural Science Foundation Projects (No.2022JJ30668 and No. 2022JJ30046).

\bibliography{refs}

\appendix
\section{Appendix}
\subsection{A. Analysis of Incremental
Weight}\label{apx.Delta_norm_analysis}

\begin{figure*}[t]
 \centering
\includegraphics[scale=0.8, trim=110 0 20 0,clip]{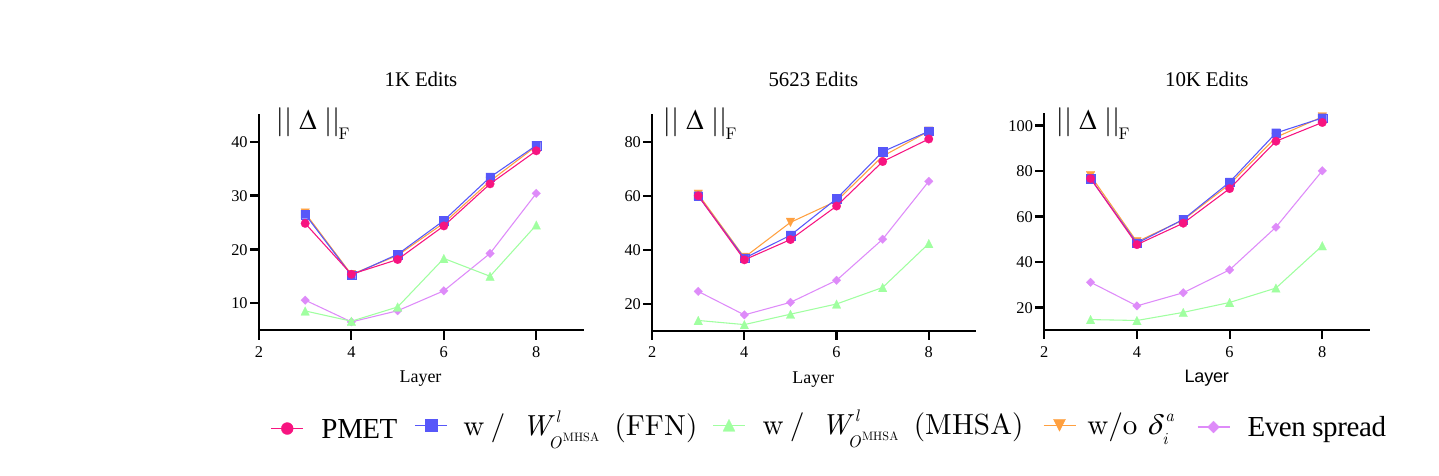}
 \caption{The norm changes of the incremental weight $\Delta$ in the ablation experiment.}\label{pmet.delta_norms}
\end{figure*}

In the ablation experiments discussed in Section \ref{pmet.sec.ablation}, we have demonstrated that PMET strikes a good balance between reliability and specificity. In this section, we further analyze the relationship between the editing performance and the norms of the incremental weight obtained in the corresponding ablation experiments. The changes in the incremental weight norms for each case are illustrated in Figure \ref{pmet.delta_norms}.

Firstly, it is evident that the norms of incremental weight for updating the MHSA weights (i.e., $W^l_{O^{\text{MHSA}}}$ (MHSA)) and the even spread case are significantly smaller than those for the other cases. In GPT-J, the the MHSA weights are 0.25 times the FFN weights, so the smaller norms for the MHSA weights are expected. The reason for the smaller norms in the even spread case is that the magnitude of the residuals $R$ is reduced more in this case compared to the square root spread case, resulting in the loss of a considerable amount of update information during the spreads of residuals, thereby compromising the reliability of the editing process. However, this information loss favors the preservation of the original model.

Secondly, the incremental weight norms for PMET, w/o $\delta^a_i$, and w/ $W^l_{O^{\text{MHSA}}}$ (FFN) are very close to each other, consistent with the findings presented in Table \ref{pmet.ablation}, where these three cases exhibit similar editing performance. A closer examination reveals that the incremental weight norm for PMET is the smallest among the three cases, but at the same time, PMET achieves the best overall performance. This implies that the incremental weight obtained in PMET are the most accurate in capturing the updates needed for the desired editing.
\subsection{B. Metrics of Model Editing Problem}
\label{apx.Metrics_of_Model_Editing_Problem}
Based on the subject-centric model editing problem, two evaluation metrics naturally emerge: reliability and specificity. Reliability aims to assess the success rate of the edited model on all knowledge related to the target knowledge, while specificity, also known as locality \cite{mitchell2022memorybased}, aims to evaluate the success rate of the edited model on knowledge sets unrelated to the target knowledge. However, evaluating the success on all knowledge related or unrelated to the target knowledge is currently challenging. Previous works \cite{Meng2022Locating,meng2022massediting} divided reliability into efficacy and generalization, evaluating the success rate of the edited model on edit sequences and paraphrase edit sequences, respectively. The paraphrase edit sequences and edit sequences are semantically consistent, while differing in syntax.

To adapt the definition of subject-centric model editing problem to existing metrics and datasets, we divide the knowledge set related to subject $S$ into explicit and implicit knowledge sets:
\begin{equation}\label{pmet.knowledge_set_ei}
\begin{aligned} \scriptsize
    K^{S_t^{\text{exp.}}}  &= \left\{ \left\{ {\boldsymbol{x}}^{S_t^{\text{exp.}}}_q,{\boldsymbol{y}}^{S_t^{\text{exp.}}}_q \right\}, q \in  \left\{0,1,2,...,Q\right\} \right\}\\
    K^{S_t^{\text{imp.}}}  &= \left\{ \left\{ {\boldsymbol{x}}^{S_t^{\text{imp.}}}_p,{\boldsymbol{y}}^{S_t^{\text{imp.}}}_p \right\}, p \in  \left\{0,1,2,...,P\right\} \right\}
\end{aligned}
\end{equation}where $P+Q= N'$ and $P \gg Q$. Explicit knowledge set consists of knowledge clue and knowledge point pairs directly shown in the text, while implicit knowledge set contains a large number of knowledge derived from the corresponding explicit knowledge. For instance, for the subject `Forbidden City,' an explicit knowledge clue may be: ``Forbidden City is located in,'' with explicit knowledge point: ``Beijing.'' An implicit knowledge clue corresponding to this explicit knowledge could be: ``If you want to go to Forbidden City, you need to take a plane from New York to,'' with the implicit knowledge point: ``Beijing.'' Another implicit knowledge clue could be: ``The place where Forbidden City is located in is China's'', with the knowledge point: ``capital.'' The explicit knowledge set and implicit knowledge set correspond to existing edit sequences and paraphrase edit sequences, respectively. Formally, efficacy, generalization, and specificity can be represented as:
\begin{itemize}
  \item Efficacy:
\begin{equation}\label{pmet.Efficacy}
  \scriptsize
    \mathbb{ E}_{\left\{ {\boldsymbol{x}}^{S_t^{\text{exp.}}}_q,{\boldsymbol{y}}^{S^{\text{exp.}}_{t}}_q \right\} \in K^{S^{\text{exp.}}_t}}\boldsymbol{1}[\mathbb{P}_{\mathcal{F}_{\theta^*}}({\boldsymbol{y}_q^{S_t^{\text{exp.}}}}| {\boldsymbol{x}}^{S_t^{\text{exp.}}}_q)>\mathbb{P}_{\mathcal{F}_{\theta^*}}({\boldsymbol{y}_q^{S^{\text{exp.}}}}| {\boldsymbol{x}}^{S_t^{\text{exp.}}}_q)]
  \end{equation}
  \item Generalization:
\begin{equation}\label{pmet.Generalization}
     \scriptsize
    \mathbb{ E}_{\left\{ {\boldsymbol{x}}^{S_t^{\text{imp.}}}_p,{\boldsymbol{y}}^{S^{\text{imp.}}_{t}}_p \right\} \in K^{S^{\text{imp.}}_t}}\boldsymbol{1}[\mathbb{P}_{\mathcal{F}_{\theta^*}}({\boldsymbol{y}_p^{S_t^{\text{imp.}}}}| {\boldsymbol{x}}^{S_t^{\text{imp.}}}_p)>\mathbb{P}_{\mathcal{F}_{\theta^*}}({\boldsymbol{y}_p^{S^{\text{imp.}}}}| {\boldsymbol{x}}^{S_t^{\text{imp.}}}_p)]
  \end{equation}
  \item Specificity:
\begin{equation}\label{pmet.locality}
     \scriptsize
    \mathbb{ E}_{\left\{ {\boldsymbol{x}}^{\neg S_t}_j,{\boldsymbol{y}}^{\neg S}_j \right\} \in K^{\neg S_t}}\boldsymbol{1}[\mathbb{P}_{\mathcal{F}_{\theta^*}}({\boldsymbol{y}_j^{\neg S}}| {\boldsymbol{x}}^{\neg S_t}_j)>\mathbb{P}_{\mathcal{F}_{\theta^*}}({\boldsymbol{y}_j^{\neg S_t}}| {\boldsymbol{x}}^{\neg S_t}_j)]
  \end{equation}
\end{itemize}

Additionally, Meng et al. \cite{Meng2022Locating,meng2022massediting} also employed the metrics of fluency and consistency to assess the generation capability of the edited models, and our work also takes these two metrics into consideration.
\subsection{C. Datasets Detail}\label{apx.dataset_detail}

\begin{figure}[t]
  \centering
  \includegraphics[width=0.4\textwidth]{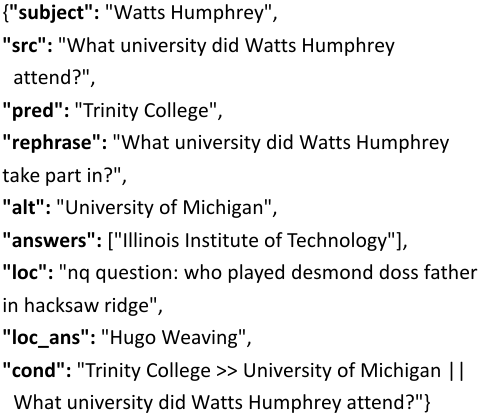}
  \caption{A sample of the zsRE dataset}\label{pmet.zsre}
\end{figure}

For the sake of fairness in testing, we use the same datasets as in MEMIT \cite{meng2022massediting}, which include zsRE and \textsc{CounterFact}. zsRE is a question-answering dataset used to evaluate the correction ability of editing methods. For example, as shown in Figure \ref{pmet.zsre}, the goal of the editing method is to change the knowledge about the subject ``Watts Humphrey'' in LLMs from ``Trinity College'' to ``University of Michigan,'' so that the edited model can answer the explicit question ``src'' (explicit knowledge) and the implicit question ``rephrase'' (implicit knowledge) correctly about this knowledge, without affecting the answer to ``loc.'' Table \ref{pmet.table-zsre} reports efficacy, which reflects the success rate of the edited model in answering explicit questions, generalization, which reflects the success rate in answering implicit questions, and specificity, which reflects the success rate in answering ``loc.'' The \textsc{CounterFact} dataset has similar testing benchmarks to zsRE for the first three items. However, \textsc{CounterFact} has more than one ``paraphrase,'' and ``loc'' (i.e., "neighborhood\_prompts" in \textsc{CounterFact}). For more details, please refer to \textsc{CounterFact} datasets \cite{Meng2022Locating}. Additionally, the \textsc{CounterFact} dataset also contains "generation\_prompts," a set of prompts containing subjects, which is used to evaluate the generation capability of the edited model, corresponding to the fluency and consistency metrics.

\subsection{D. Experimental Detail}\label{apx.experimental_detail}
The critical layers for GPT-J and GPT-NeoX have been identified as {$\mathcal{R}=\left\{3,4,5,6,7,8\right\}$} and {$\mathcal{R}=\left\{6,7,8,9,10\right\}$} \cite{Meng2022Locating,meng2022massediting}. Therefore, we mainly update these critical layers of GPT-J and GPT-NeoX. All the baselines we compare, including the parameter settings of MEMIT, are consistent with \cite{meng2022massediting}.

For the optimization of the TC hidden states in GPT-J and GPT-NeoX, we initially set $\varphi=1$ and $0 \leq \mu \leq 1$ (As $\mu$ increases, the degree of retention of the model's original knowledge becomes higher, while $\varphi$ exhibits the opposite trend.) When we have maximized the probability of the target knowledge, we want to preserve the model's original knowledge as much as possible, so we set $\varphi=0.1$ and stop the optimization when {$D_ {\text{KL}}\left(\mathbb{P}_{\mathcal{F_\theta}\left( a^L_i+=\delta^a_i, m^L_i=v^m_i\right)}\left[\boldsymbol{y} \mid p'  \right]  \lVert \mathbb{P}_{\mathcal{F_\theta}}\left[\boldsymbol{y} \mid p'  \right]  \right) <0.01$}.

On GPT-J, for the covariance matrix (i.e.,  the set of previously memorized keys $C_0$) estimation, we sampled 10K times on Wikitext in $\text{fp}32$ precision and set $\lambda=6000$ ($\lambda=4500$ in zsRE). When optimizing the TC hidden states $\delta^a_i$ and $\delta^m_i$, we set the total optimization steps to 30 with a learning rate of 0.2 (20 steps with a learning rate of 0.5 in zsRE), and limit them, as in MEMIT, to have their norms less than $\frac{3}{4}$ of the norms of the original intermediate states.

On GPT-NeoX, we sampled 5K times on Wikitext in $\text{fp}16$ precision and stored the covariance matrix in $\text{fp}32$ precision, with $\lambda=15000$. For optimization, we set the total optimization steps to 30 steps with a learning rate of 0.5, and limit the TC hidden states to have their norms less than $\frac{4}{5}$ of the norms of the original intermediate states.

As the algorithmic steps of PMET are fundamentally similar to MEMIT, the time consumption of the two methods is almost equivalent.

\end{document}